\documentclass[journal]{IEEEtran}
\usepackage{hyperref}
\usepackage[cmex10]{amsmath}
\usepackage{fixltx2e}
\usepackage{cite}

\usepackage{subcaption}
\usepackage{graphicx}
\usepackage{placeins}

\DeclareMathOperator*{\argmax}{arg\,max}

\begin{document}

\title{maskSLIC: Regional Superpixel Generation with Application to Local Pathology Characterisation in Medical Images}

\author{Benjamin~Irving, 
Iulia~A.~Popescu,
Russell~Bates,
P.~Danny~Allen, 
Ana~L.~Gomes,
Pavitra~Kannan,
Paul~Kinchesh,
Stuart~Gilchrist,
Veerle~Kersemans,
Sean~Smart,
Julia~A.~Schnabel, 
Sir~J.~Michael Brady and
Michael~A.~Chappell%
\IEEEcompsocitemizethanks{
\IEEEcompsocthanksitem B. Irving, I. Popescu, R. Bates, and M. Chappell are with the Institute of Biomedical Engineering, Department of Engineering Science, University of Oxford, UK.
\IEEEcompsocthanksitem J.A. Schnabel is with Division of Imaging Sciences and Biomedical Engineering, King's College London, UK.
\IEEEcompsocthanksitem Sir J.M. Brady, A. Gomes, D. Allen, P. Kannan, P. Kinchesh, S. Gilchrist, V. Kersemans and S. Smart are with the Department of Oncology, University of Oxford, UK.
}}%

\maketitle
\IEEEtitleabstractindextext{%
\begin{abstract}
Supervoxel methods such as Simple Linear Iterative Clustering (SLIC) are an effective technique for partitioning an image or volume into locally similar regions, and are a common building block for the development of detection, segmentation and analysis methods. We introduce \emph{maskSLIC} an extension of SLIC to create supervoxels within regions-of-interest, and demonstrate, on examples from 2-dimensions to 4-dimensions, that maskSLIC overcomes issues that affect SLIC within an irregular mask. We highlight the benefits of this method through examples, and show that it is able to better represent underlying tumour subregions and achieves significantly better results than SLIC on the BRATS 2013 brain tumour challenge data (p=0.001) -- outperforming SLIC on 18/20 scans.  Finally, we show an application of this method for the analysis of functional tumour subregions and demonstrate that it is more effective than voxel clustering. 
\end{abstract}
}

\IEEEdisplaynontitleabstractindextext
\IEEEpeerreviewmaketitle

\section{Introduction}

Superpixel/voxel methods partition an image or volume into local meaningful subregions \cite{achanta2012slic,Vedaldi2008,Felzenszwalb2004}. These methods capture local similarity while at the same time reducing the redundancy of the images, speeding up processing, and, therefore, making complex analysis of regional relationships more feasible. An additional benefit is that these methods move beyond a rigid pixel/voxel structure to a representation that is more robust to noise and partial voluming.  

Superpixels (and the extension to 3D volumetric images known supervoxels) have seen widespread application including object detection in computer vision \cite{fulkerson2009cso}, segmentation in microscopy and anatomical images \cite{mahapatra2013ads,lucci2012} and  extended to perfusion images such as dynamic contrast enhanced magnetic resonance imaging (DCE-MRI) for automated tumour segmentation \cite{Irving2014act,Irving2016pieces}. Simple linear iterative clustering (SLIC) \cite{achanta2012slic} has been shown to be a fast and effective method of generating superpixels/supervoxels. Recently, SLIC has been used for the subregional assessment of tumours \cite{Conze2016} and organs such as the heart \cite{Popescu2016}. 

In 3D volumetric scans, the supervoxel representation has considerable potential for capturing subregional heterogeneity in areas of pathology such as a tumour. However, supervoxel methods are generally designed for whole image analysis, which results in problem-specific challenges when subregions are assigned within an irregular region, such as a tissue specific mask, which is common in medical imaging applications. 


In this study we use a variety of illustrative examples and real tumour imaging to illustrate the need for a masked supervoxel approach for 2D, 3D and 4D volumes. We use the term 4D volumes to describe volumes that include the temporal domain; in DCE-MRI volumes a number of volumes are acquired over time to capture changes in perfusion. As the main application of this method is analysis of pathology such as real tumours, in Section \ref{sec:back} we introduce current approaches to tumour subregional analysis. We then extend  (SLIC) from an image-based method to a method that is specific to irregular regions-of-interest (or masks), which we call \textit{maskSLIC} (Section \ref{sec:slic}). In Section \ref{sec:app} we compare \emph{maskSLIC} to SLIC and demonstrate its application on both medical and non-medical examples.   

\section{Background on tumour analysis} \label{sec:back}

Tumour subregional analysis is becoming an important component of tumour assessment \cite{OConnor2014} and is one of the motivations for a new method that works within an irregular mask. We use examples of tumour subregional analysis to demonstrate our method, and thus introduce some key concepts to tumour growth and imaging here.  

Angiogenesis, the formation of new vessels, plays an important role in tumour progression and tumours exhibit chaotic and leaky vasculature which leads to poorly and well perfused regions. This prompts the formation of regions of edema, hypoxia and necrosis. Contrast-enhanced imaging such as dynamic contrast-enhanced magnetic resonance imaging (DCE-MRI) or static contrast-enhanced T1-weighted MRI provide a way of assessing tumour perfusion. In a recent review, O'Connor \emph{et al.} (2014) \cite{OConnor2014} highlight the identification of tumour subregions that define local tumour biology as a key method to quantify tumour heterogeneity. To achieve this, a number of methods use either location based definitions such as radial subregions for a sphere-like tumour, or manually set thresholds on imaging parameters \cite{OConnor2014}. 

These methods rely on rigid definitions of tumour regions, and as an alternative, clustering based on imaging parameters is often used as an unsupervised and data-driven approach of defining tumour regions \cite{OConnor2014,Castellani2009}. Clustering is dependent on the distribution of image derived features within the tumour and, therefore, the region centroids may vary on a case by case basis. For stability and generalisability, the clustering should, therefore, be performed jointly on a collection of cases as demonstrated by Henning et al (2007) \cite{Henning2007}; otherwise, the region definitions will vary between cases. However, a voxelwise clustering across a large number of cases leads to a very large number of data points based on potentially noisy, motion affected or partially volumed parameter maps without any local spatial regularisation.
 
Supervoxelisation can be a useful step in the extraction of spatially regularised tumour subregions that reduce noise while allowing regional comparison across an entire dataset. The \emph{mask}SLIC supervoxel-based analysis method that we propose has several advantages over voxelwise approaches for many problems involving the subregional assessment of tumours, including: 1) providing spatial regularisation to reduce noisy voxel outliers, 2) reducing the representation of each tumour to \emph{n} regions which allows efficient comparison across the large datasets of cases, and 3) providing a scale invariant representation that is independent of tumour size. 1) also applies to standard supervoxel methods such as SLIC but as demonstrated in the following sections, SLIC is not suited to irregular regions and the number of supervoxels inside an irregular region cannot be set.  

\section{maskSLIC method} \label{sec:slic}

The original SLIC method \cite{achanta2012slic} is initialised using a grid of cluster centres that are placed equidistantly on the image. A local k-means clustering is applied to assign each voxel to a cluster centre. The cluster centres are updated based on the assigned voxels and the process is iterated. The distance function ($d$), used in the clustering, combines the spatial and feature similarity as shown in Eqn \ref{eqn:dist}, where $r$ is a weight between the feature distance $d_f$ and the spatial distance $d_s$. 


\begin{align}
d = \sqrt{(d_f)^2 + \left(d_s/r\right)^2} \label{eqn:dist}
\end{align}

SLIC provides an efficient coding of the image for tasks such as recognition and segmentation, but when defining subregions inside a mask, such as an organ or tumour, this definition can be problematic because the method is initialised from a grid of seed-points. 

Two naive approaches to using SLIC with a mask are as follows: 
\begin{enumerate}
\item Apply SLIC to the entire image or volume and then identify supervoxels contained in the mask (for partially overlapping supervoxels, only the region contained in the mask is kept) 
\item Alternatively, after grid initialisation, keep seed points that fall inside the mask and apply SLIC only to voxels inside the mask. 
\end{enumerate}

\begin{figure}[t]
\centering
\includegraphics[width=7cm]{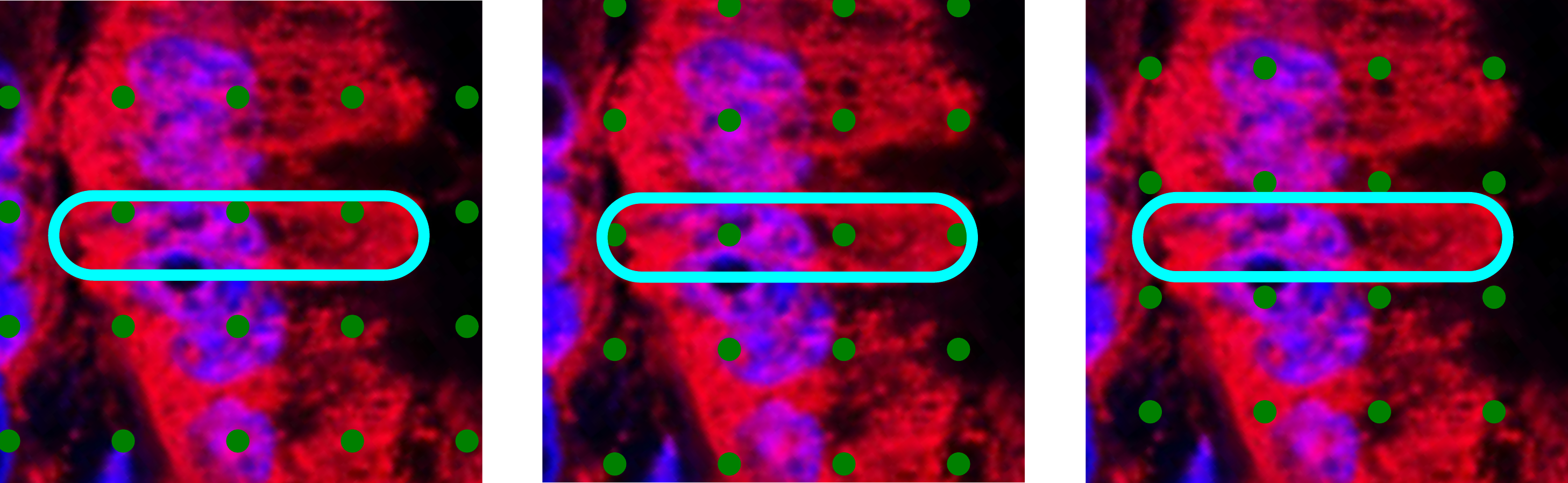}
\caption{An illustrative example showing grid seed points initialised by SLIC (green) and a small mask (turquoise). Depending on the location of the grid relative to the mask, the mask could contain 3 seed points on the edge of the region, 4 seed points, or zero seed points. } \label{fig:toy}
\end{figure}

Choosing 1 means that supervoxels in the mask are affected by the surrounding image variation, which may lead to supervoxels that are only partially in the region. Choosing 2, and only considering voxels within the mask, would seem a better choice but is dependent on where the seed points are placed with respect to the mask, which is illustrated in the illustrative example with three identical masks (Figure \ref{fig:toy}). One mask happens to contain four seed points, while the others three and zero, respectively, leading to failure of the supervoxel method in the third case. This example also highlights why neither methods 1 or 2 are translation invariant.


\begin{figure*}[htp]
\centering
\includegraphics[width=16cm]{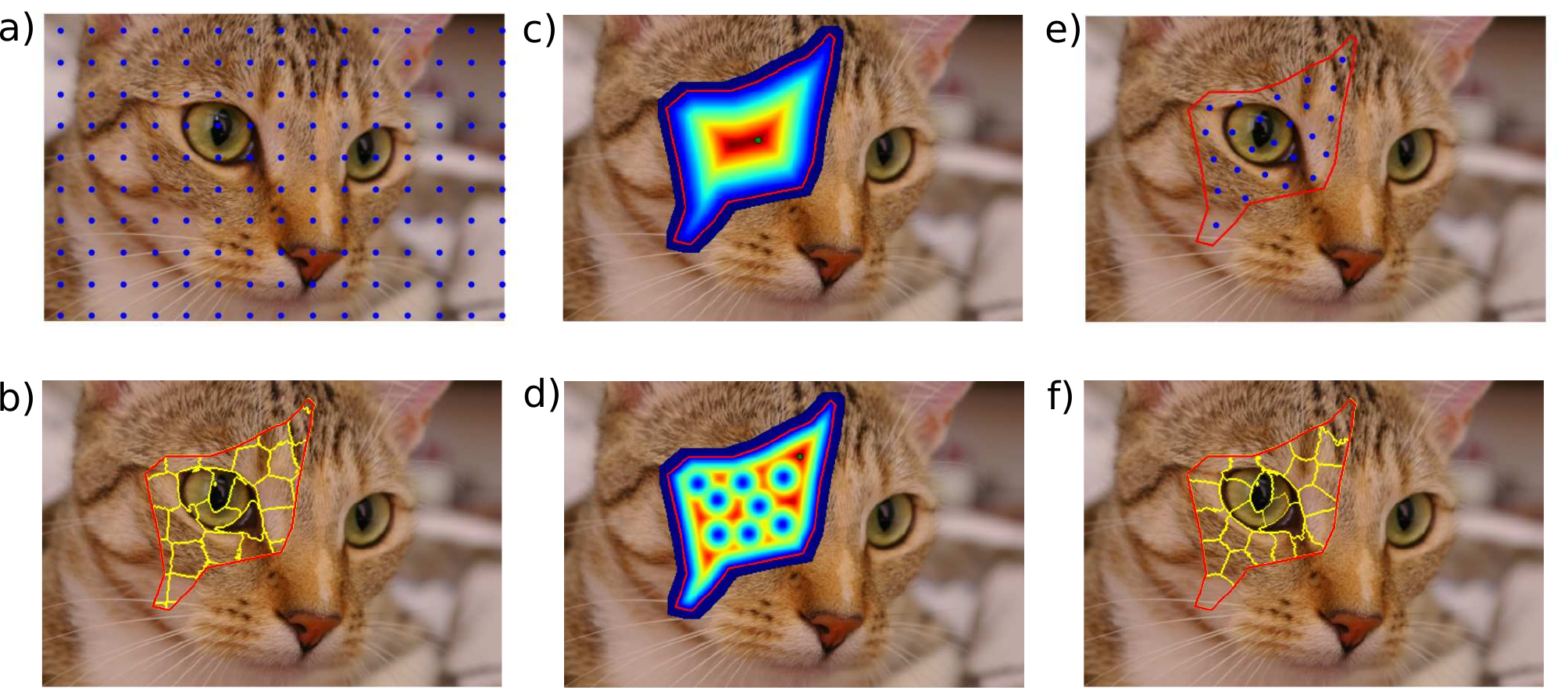}
\caption{Initialising SLIC seed points inside a mask.  a), b) show the original SLIC formulation. c), d) Show the distance transform heat map (red=high, blue=low) for iteratively placing each seed points, e) shows the final seed points, and f) shows the final \emph{mask}SLIC supervoxels.  \emph{mask}SLIC (f) produces supervoxels that are consistent and regular with respect to the mask while SLIC (b) produces small and irregular supervoxels at the region border, with some supervoxels influenced strongly by the surrounding image. \emph{Image is CC0 with no copyright restrictions from photographer Stefan van der Walt}.} \label{fig:slic}
\end{figure*}

Since both naive approaches are inconsistent and can lead to failure of the SLIC method for masks, we need an approach where the subregions are more robustly generated with respect to the shape of the mask. We propose a new method called \emph{mask}SLIC, which makes three modifications to the original method and is shown in Figure \ref{fig:slic}. Two steps are used to generate improved initialisation points within a mask, and, once the points are initialised, the clustering is limited to voxels inside the mask -- as a third step. These are detailed as follows:

\textbf{Step 1}) Given a specified number of supervoxels (\emph{N}), a Euclidean distance transform is used to iteratively place seed points spaced at the maximum distance from the boundaries and any other seed points (Figure \ref{fig:slic} c and d). Given a mask, $B$ is the set of background (non-mask) labels and $L$ is the set containing background (B) and labelled points (P) i.e. $ L = B \cup P $; initially $P = \{\}$. 


$D(\boldsymbol{x})$ is the distance transform at location $\boldsymbol{x}$:

\begin{equation}
D(\boldsymbol{x}) = \min_{y \in L} \left(\sum^n_i (x_i-y_i)^2\right)^{\frac{1}{2}}
\end{equation}
where $n$ is the number of spatial dimensions. The furthest distance $p^*$ is found:

\begin{equation}
p^* = \argmax_x D(\boldsymbol{x})
\end{equation}

and $P$ becomes $P \cup \{p^*\}$ for the next iteration which is repeated $N$ times.

\textbf{Step 2}) SLIC is applied to the seed points ($P$) with only the distance feature ($d_s$ from Equation \ref{eqn:dist}): 

\begin{equation}
d = d_s
\end{equation}

This acts to optimise the location of the seed points given the shape of the mask based on a k-means distance metric from the initialised seed points (Figure \ref{fig:slic}e).

\textbf{Step 3}) Finally, SLIC is applied to voxels that are defined inside the mask (figure \ref{fig:slic}f).
 
The first step provides a mechanism to spatially distribute seed points within a mask and the second step promotes even placement of the points. Figures \ref{fig:slic} a) and b) show the SLIC grid initialisation points which are unevenly distributed within the mask, which consequently affects supervoxel generation, particularly when the region border is poorly defined. Figures \ref{fig:slic} e) and f) show the proposed \emph{maskSLIC} method.  A demonstration of our \emph{mask}SLIC method is available at: \href{http://maskslic.birving.com}{\underline{http://maskslic.birving.com}}.

Our method amounts to using an implicit coordinate frame based on the mask instead of an extrinsic coordinate frame used in SLIC. The seeding method shares some of the goals of \emph{k-means++} \cite{arthur2007k} for improved and robust seeding of k-means, but in this case the region is explicitly defined. Importantly, by doing so it is possible to allow the user to specify the number of required supervoxels inside the mask, which cannot be done with the SLIC method.

\begin{figure*}[htp]
\centering
\includegraphics[width=17cm]{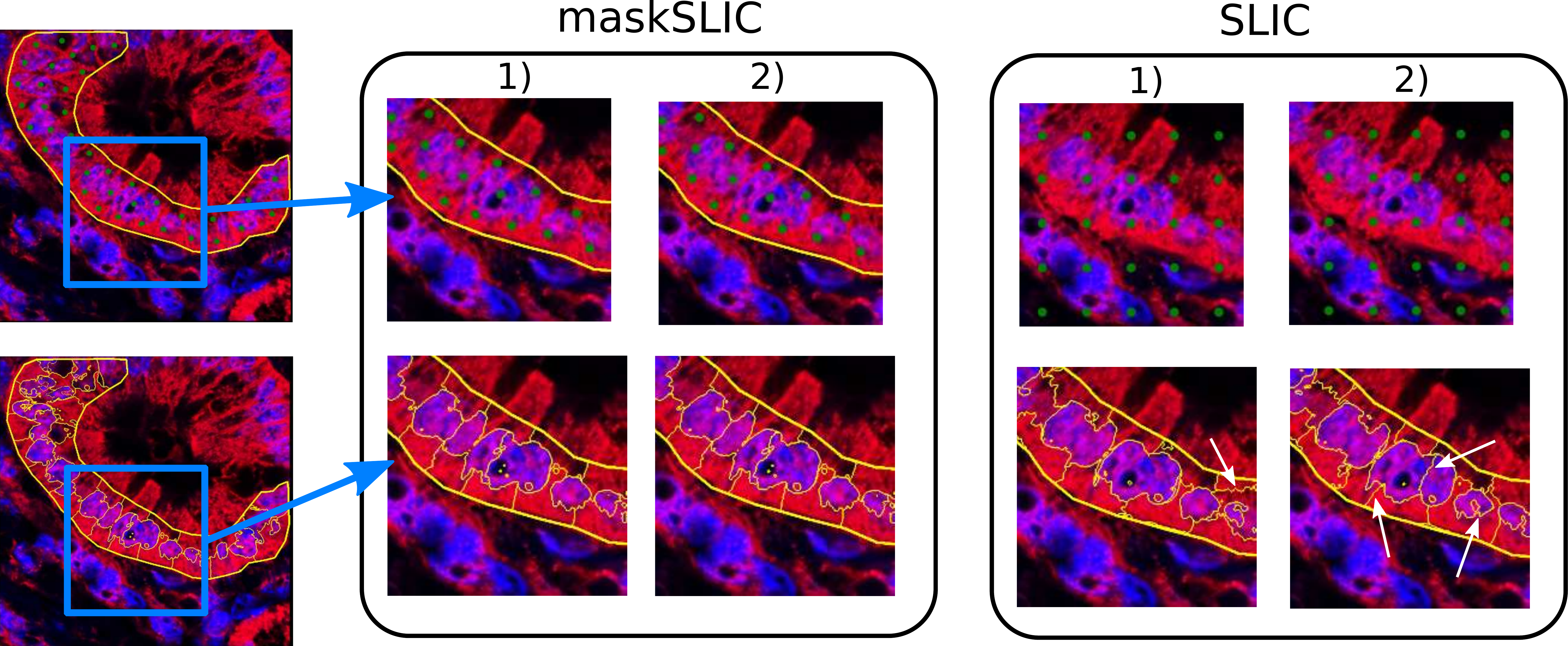}
\caption{Effect of seed points on supervoxel subregion generation. \emph{maskSLIC} is compared to SLIC for the initial location (1) and after a translation of 40 pixels (2). \emph{mask}SLIC regions are identical because the generation is defined on the mask, while the superpixel representaton of the regions in SLIC are clearly affected by the translation. White arrows illustrate some of the differences in the superpixel boundary definitions in SLIC. This example image shows mitochondiral (red) and nuclear (blue) staining of a pancreatic carcinoma (The image is in the public domain, David Kashatus, National Cancer Institute)} \label{fig:trans1}
\end{figure*}

\section{Results} \label{sec:app}

In this section we first assess the effect of seed point initialisation on supervoxels generated using SLIC and \emph{mask}SLIC. Next we evaluate the effectiveness of the two methods to represent underlying tissue on the BRATS 2013 dataset. Finally, we demonstrate the potential of this supervoxel approach for characterising tumour regions in 4D perfusion data. This section serves to demonstrate the general applications of our proposed method ranging from 2D to 4D, the benefits over SLIC for regional analysis, and its application to tumour imaging. 

\subsection{Example of translation invariance in 2D}

In this first assessment of maskSLIC we define a mask on an image to demonstrate the translation invariance of \emph{mask}SLIC. A microscopy image of a pancreatic carcinoma, made available by the National Cancer Institute, is used and an arbitrary mask is chosen that contains both mitochondrial and nuclear staining.

We examine the effect of placement of seed points with respect to the mask by translating an image and regenerating the supervoxels. Figure \ref{fig:trans1} shows the effect of seed point placement for the original SLIC method and our proposed \emph{maskSLIC} method. Each method shows the superpixel generation before (1) and after (2) a translation of 40 voxels. \emph{maskSLIC} regions are identical because the generation is defined by an implicit reference frame derived from the mask, while the supervoxel representation of the regions in the other methods are clearly affected by the translation.

\begin{figure}[t]
\centering
\includegraphics[width=6cm]{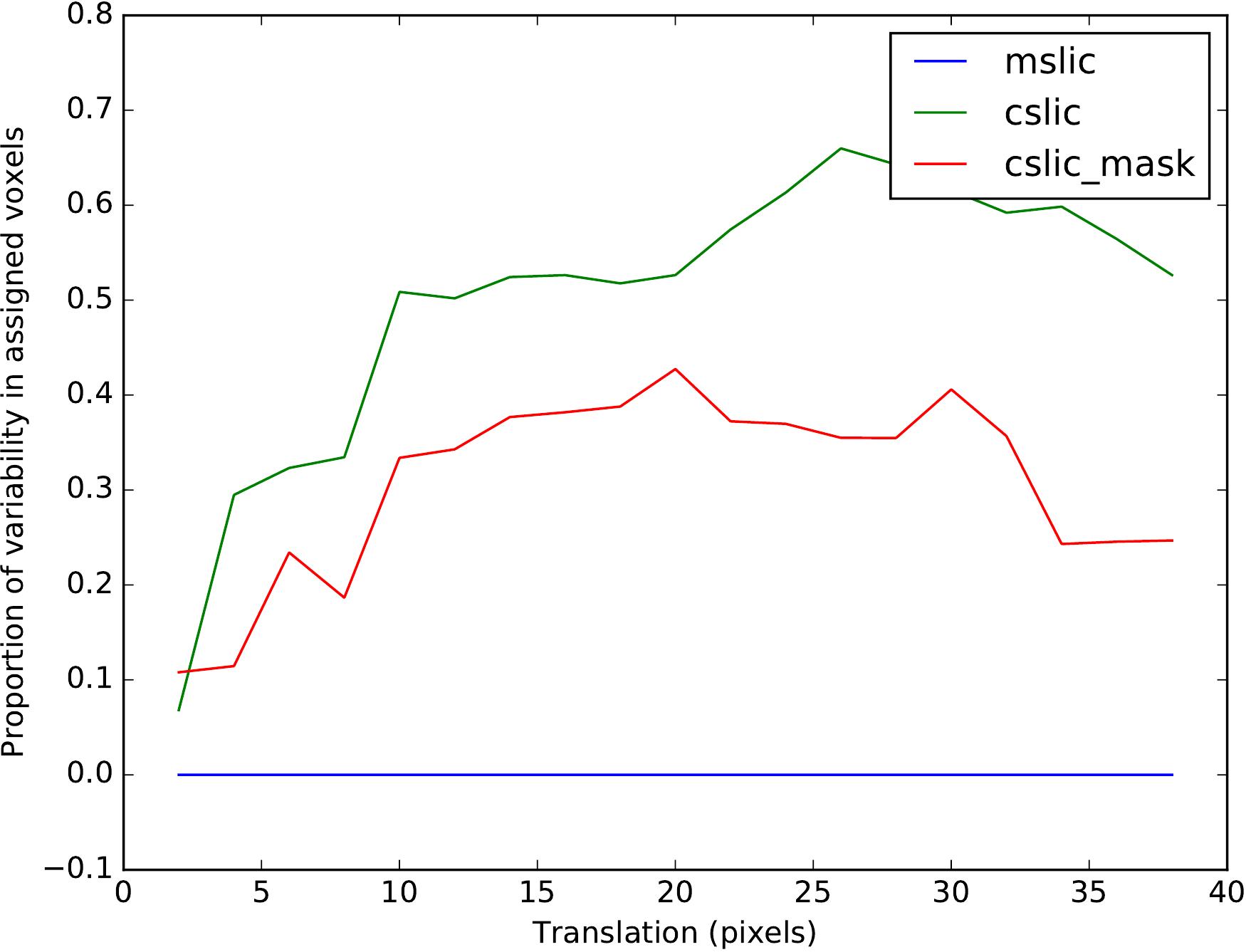}
\caption{Variation $C_s$ (Equation \ref{eq:cs}) in the supervoxel region definition as an effect of translation for \emph{mask}SLIC (mslic), SLIC (cslic) and SLIC constrained to the mask (cslic\_mask). Restricting SLIC to the mask (cslic\_mask) shows some improvement; \emph{mask}SLIC has an error of 0.} \label{fig:trans2}
\end{figure}

To quantify differences for each method when subjected to translation, we define $C_s$ as the mean of the best overlap scores between supervoxels before ($S_1$) and after translation ($S_2$):

\begin{align}
\delta_s(p) = \max_{q \in S_2}\,\mathrm{DSC}(p, q)\\
C_s = \frac{1}{N} \sum_{p \in S_1} 1- \delta_s(p) \label{eq:cs}
\end{align}

where N is the total number of supervoxels in the mask and $\delta_s(p)$ is the maximum DICE overlap between supervoxel $p \in S_1$ and any supervoxel in $S_2$.  $C_s$ provides a mechanism to assess the similarity of the supervoxels without knowing the one-to-one correspondence between two generated sets of regions. Figure \ref{fig:trans2} shows the supervoxel variation ($C_s$) with translation (our proposed method \emph{maskSLIC} has zero change).

\subsection{Quality of the representation of the underlying image}

We now demonstrate that \emph{mask}SLIC provides more meaningful subregions within a mask.  We used data from the BRATS 2013 challenge (braintumorsegmentation.org) \cite{menze2015}. This data consists of 20 pre-therapy scans of high grade glioma patients.

\begin{figure*}[htp]
\centering
\includegraphics[width=14cm]{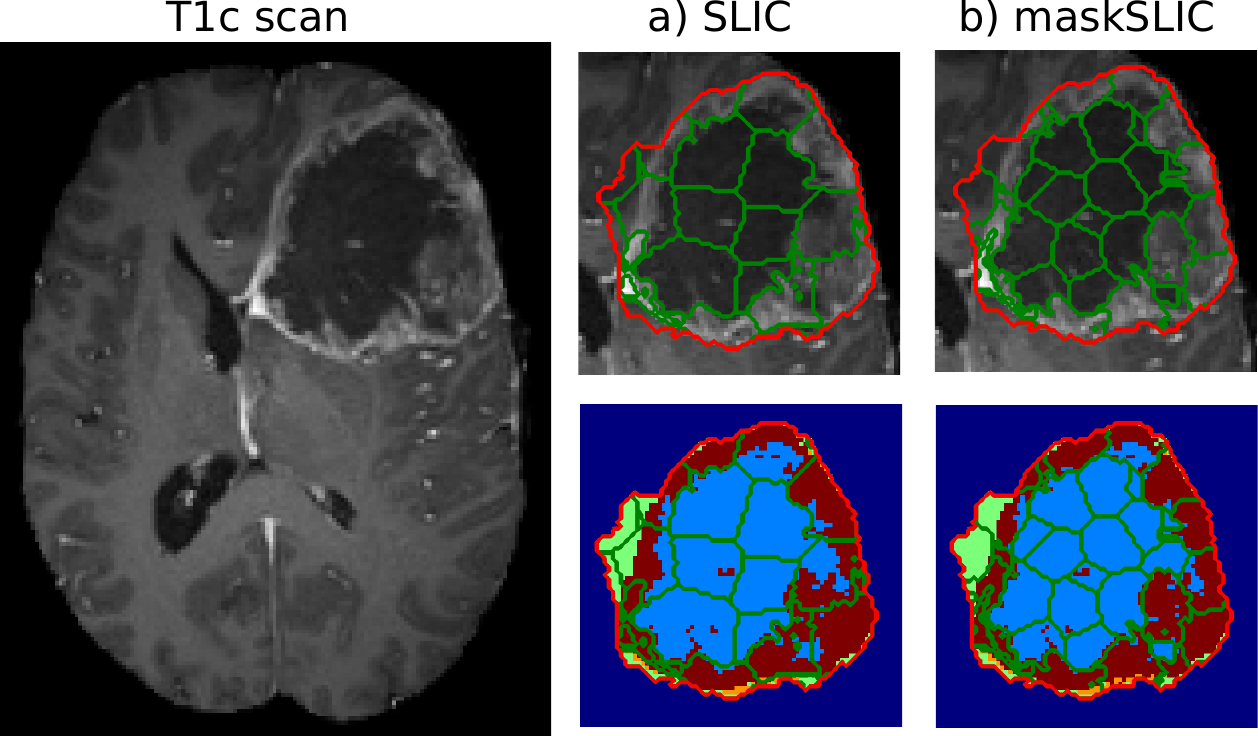}
\caption{SLIC and maskSLIC oversegmentation of a T1 contrast image of a high grade glioma from the BRATS 2015 challenge \cite{menze2015} with manual regional labels derived from  T1 and T2 images. The labels are as follows with the figure label colour where visible: 1) necrotic centre (blue), 2) edema (not visible in figure), 3) non-enhancing gross abnormalities (green) and 4) Enhancing regions with gross tumour abnormalities (red). Note that \emph{mask}SLIC appears to have a greater number of smaller supervoxels. This is because the supervoxels are better distributed throughout the volume.} \label{fig:brats}
\end{figure*}

SLIC was used to oversegment the T1 contrast-enhanced MRI scan (T1c) of a patient with a glioma tumour into 2000 supervoxel regions using approach 1 of Section \ref{sec:slic}. \emph{maskSLIC} was generated using the same number of supervoxels as found previously with SLIC (Fig \ref{fig:brats} b). Next the four ground truth tumour labels (necrotic centre, edema, non-enhancing gross abnormalities and enhancing regions) were used to assess the label consistency ($l_c$) of each subregion, which was defined as the proportion of voxels in each supervoxel that are the majority label.

\begin{figure}[t]
\centering
\includegraphics[width=8cm]{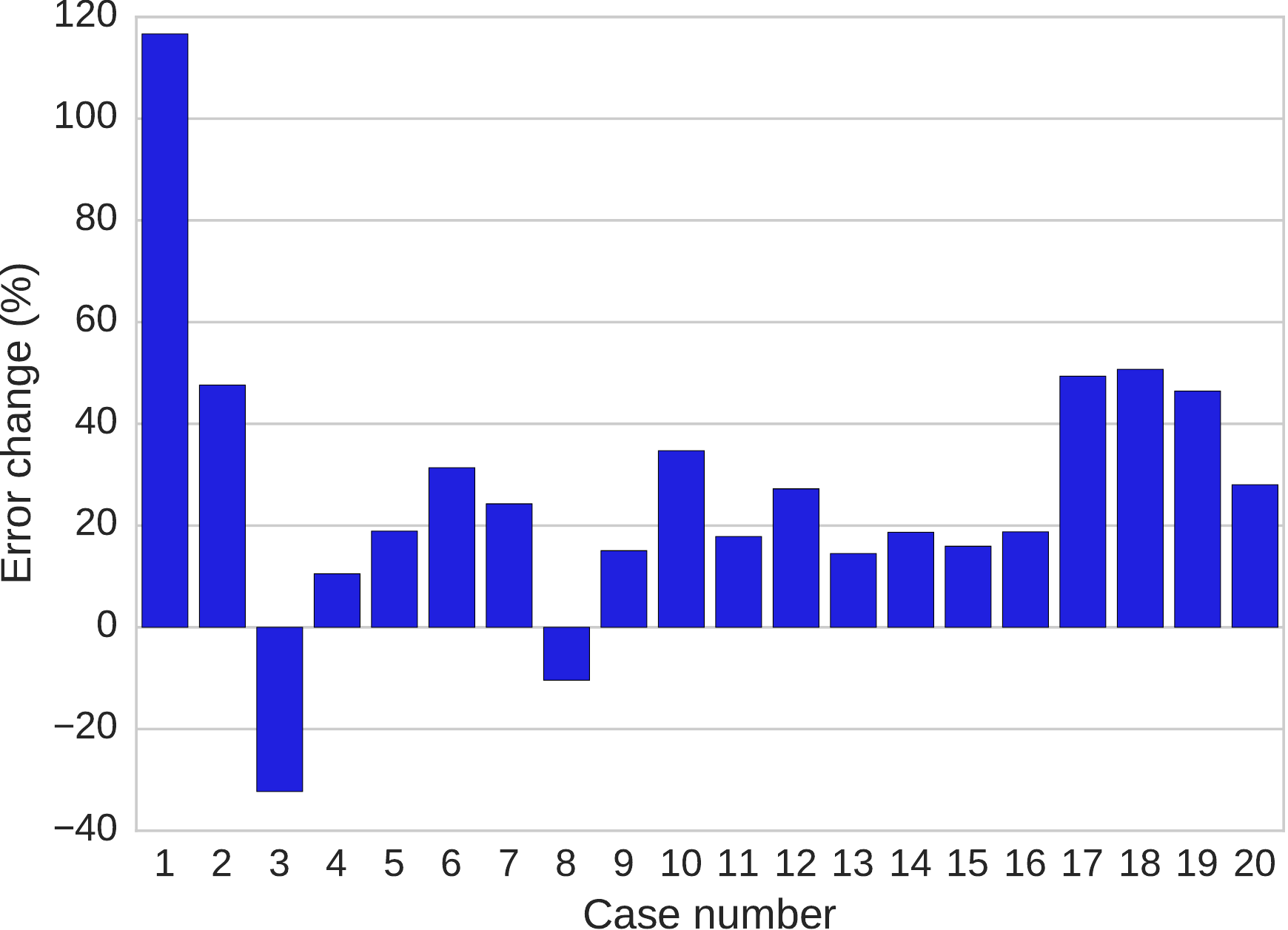}
\caption{Percentage change in the error ($E$) when using SLIC compared to \emph{mask}SLIC} \label{fig:ei}
\end{figure}

For the dataset of 20 scans, SLIC obtained a median $l_c$ of 85\% while maskSLIC obtained a median $l_c$ of 89\%. This improvement was significant (p=0.001 using Wilcoxon signed rank). The percentage error increase of using SLIC compared to maskSLIC for each case is defined as  $E =100 \left( \frac{e_{slic} - e_{mslic}}{e_{mslic}} \right)$ where the error for each method is defined as $e = 1-l_c$,  and is shown in Figure \ref{fig:ei}. maskSLIC outperformed SLIC in 18/20 cases as shown in Figure \ref{fig:ei}.

The mean time to process a single 3D scan was 21.6sec for SLIC and 14.96sec for \emph{mask}SLIC. The process of assigning seed points is slower in \emph{mask}SLIC, however since only regions inside the ROI are computed, this leads to faster processing for small regions.

In summary, for the same number of subregions, maskSLIC produces more meaningful and well distributed regions, and achieves a better label consistency and size. Note that we are using differences in the error to illustrate the differences between the methods. An overall improvement in the error could potentially be achieved for both SLIC and \emph{mask}SLIC by using larger number of regions or multiple features.

\subsection{Application to monitoring tumour cohorts} \label{sec:tumourp}

In this section, we demonstrate the use of \emph{maskSLIC} as a preprocessing step to improve unsupervised clustering of 4D DCE-MRI perfusion scans and create meaningful tumour subregions. Our aim is to provide a more robust approach than voxelwise k-means clustering based on the perfusion features, by performing the clustering on \emph{mask}SLIC supervoxels, which extends on our accepted abstract \cite{Irving2017mrs}.

\subsubsection{Experimental set-up}

To  evaluate  the  method,  we  used scans from a study that performed daily DCE-MRI preclinical tumour imaging to monitor tumour growth for 10 cases over 8 days. DCE-MRI was performed at 4.7 T (Varian VNMRS) using a cardio-respiratory gated spoiled 3D gradient  echo scan  with TE 0.64 ms, TR 1.4 ms, nominal flip angle 5 degrees at a voxel  size 0.42x0.42.0.42 $mm^3$ and 60 frames, each taking ca. 10-15 s, dependent upon the imstantaneus respiratory and heartbeat rates. Respiration was monitored using a pressure balloon, and ECG with subcutaneously implanted needles. Motion artefact was minimised  with  cardiac synchronization and the automatic and immediate reacquisition of data corrupted by  respiration motion. A Gadolinium based contrast agent was injected (30 ul over 5 s) after image 10/60  to image the perfusion  through the tumour. Quantitative tissue T1 values were determined prior to DCE-MRI from a variable flip angle scan with the same CR-gated  3D gradient echo scan but with and 16 nominal FAs ranging from 1-7 degrees in steps of 0.4 degrees.

\begin{figure}[htp]
\centering
\begin{subfigure}{.2\textwidth}
\includegraphics[height=3.5cm]{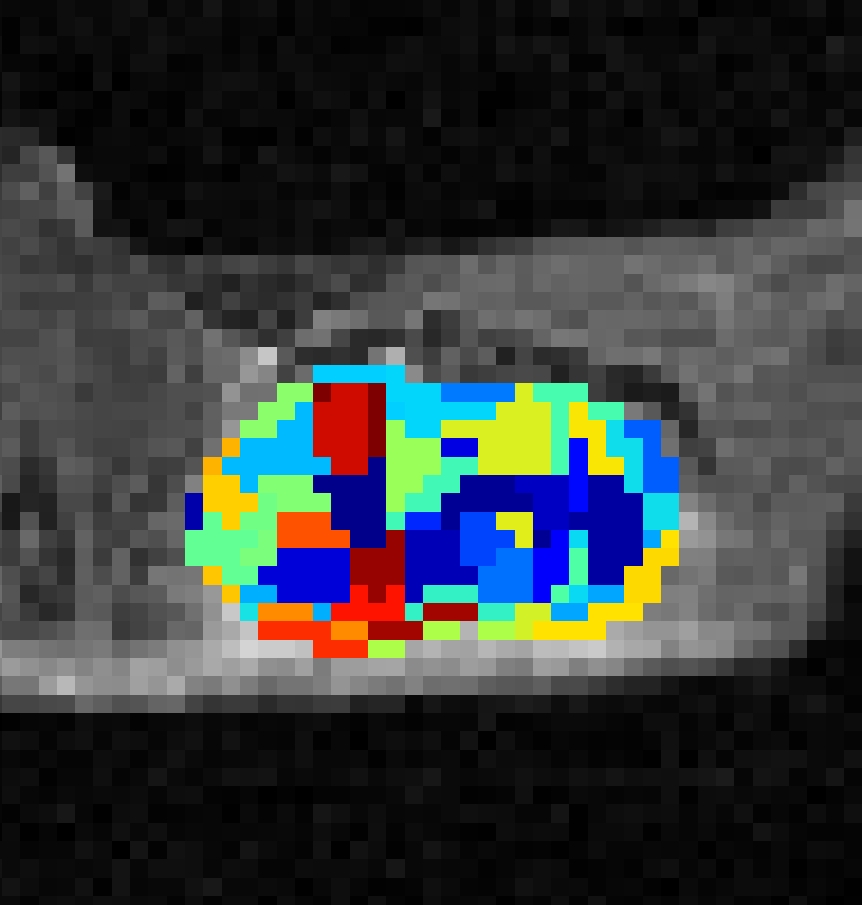}
\end{subfigure}
\begin{subfigure}{.2\textwidth}
\includegraphics[height=3.5cm]{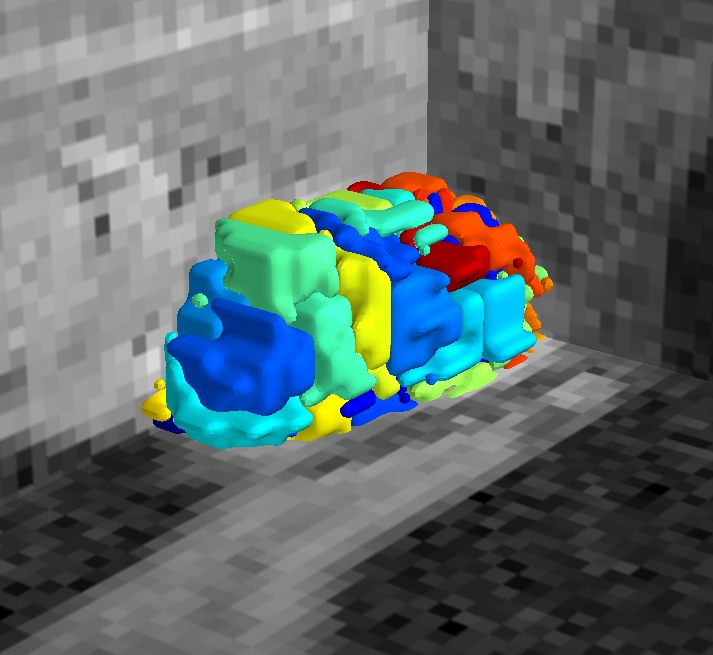}
\end{subfigure}
\caption{3D supervoxels extracted from a 4D DCE-MRI scan} \label{fig:perfsuper}
\end{figure}

We applied \emph{mask}SLIC to the 4D perfusion images using the principal component decomposition of perfusion to extract supervoxels as outlined in \cite{Irving2014act} and \cite{Irving2016pieces}. Each tumour was decomposed into spatially contiguous supervoxels with similar perfusion, as shown in Figure \ref{fig:perfsuper}. Perfusion features $K^{trans}$, $k_{ep}$ and $T_1$ were calculated from the scan using the Toft's model \cite{tofts1997}. 

Next we applied k-means clustering to both the supervoxels (our method) and image voxels (standard approach) using the perfusion features. In both cases all supervoxels or voxels from the entire dataset were clustered into four distinct labels (or regions). 

\subsubsection{Results}

\begin{figure*}[htp]
\centering
\includegraphics[width=12cm]{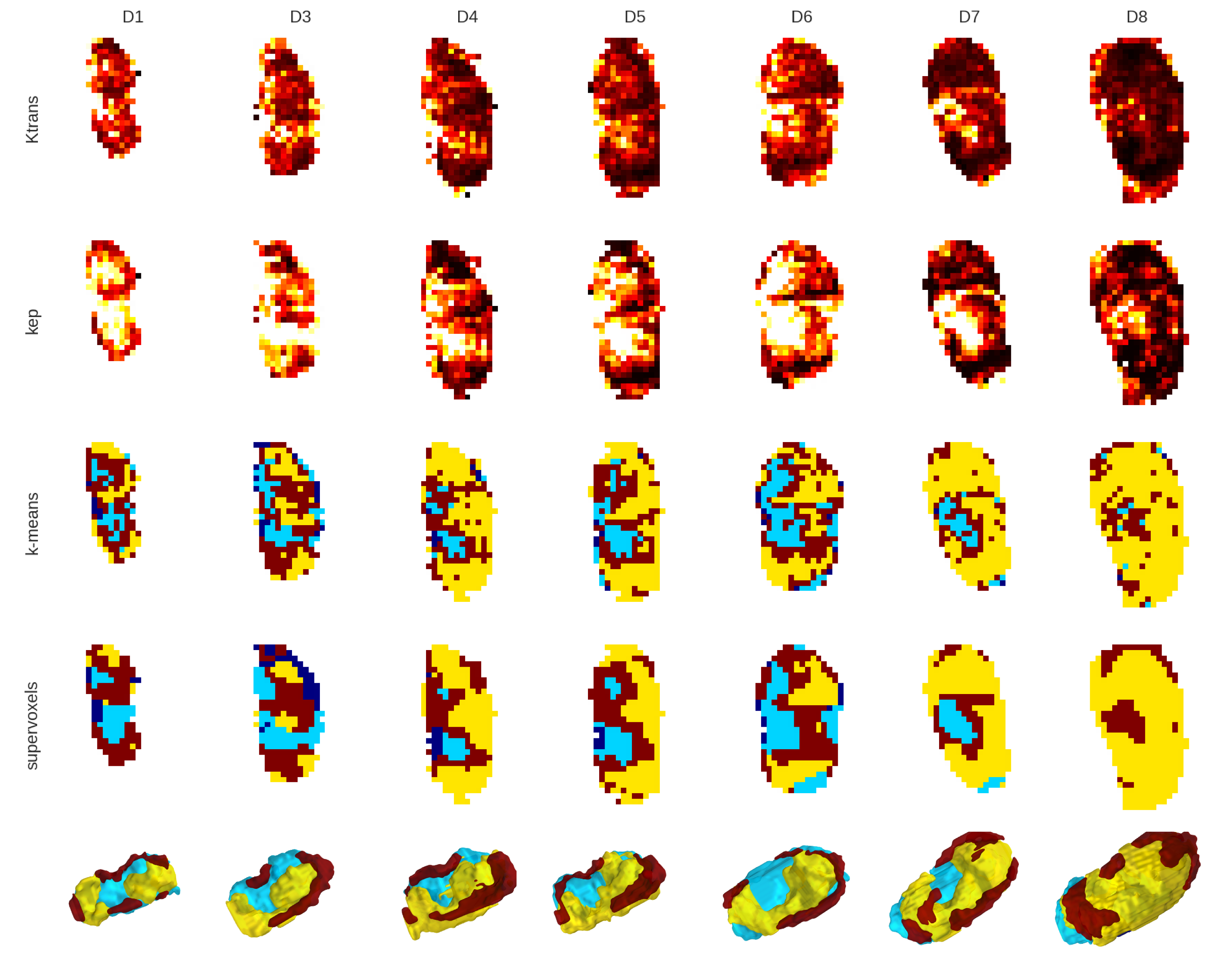} 
\caption{Trends in a tumour over 8 days, showing $K^{trans}$ and $k_{ep}$ maps, voxelwise k-means clustering, and k-means clustering with \emph{mask}SLIC supervoxel processing: D1 - D8 (D7 is not shown due to scan failure). A 3D rendering of the clustering supervoxel regions is also shown to illustrate the regional consistency during tumour growth, with the necrotic region expanding. } \label{fig:reg1}
\end{figure*}

Figure \ref{fig:reg1} shows the progression of a single tumour over 8 days in terms of $K^{trans}$, $k_{ep}$ and labelled supervoxel regions. Voxelwise k-means clustering was noisy, while clustering supervoxels using \emph{maskSLIC} provided well-defined subregions. The better definition provided by \emph{mask}SLIC facilitated 3D rendering of the tumour regions, as shown in Figure \ref{fig:reg1}, allowing the trends in the subregions to be visualised more readily over time. These regions also show consistency (with changes due to growth) over multiple acquisitions indicating that these truly represent biologically distinct regions. 

\begin{figure*}[htp]
\centering
\includegraphics[width=14cm]{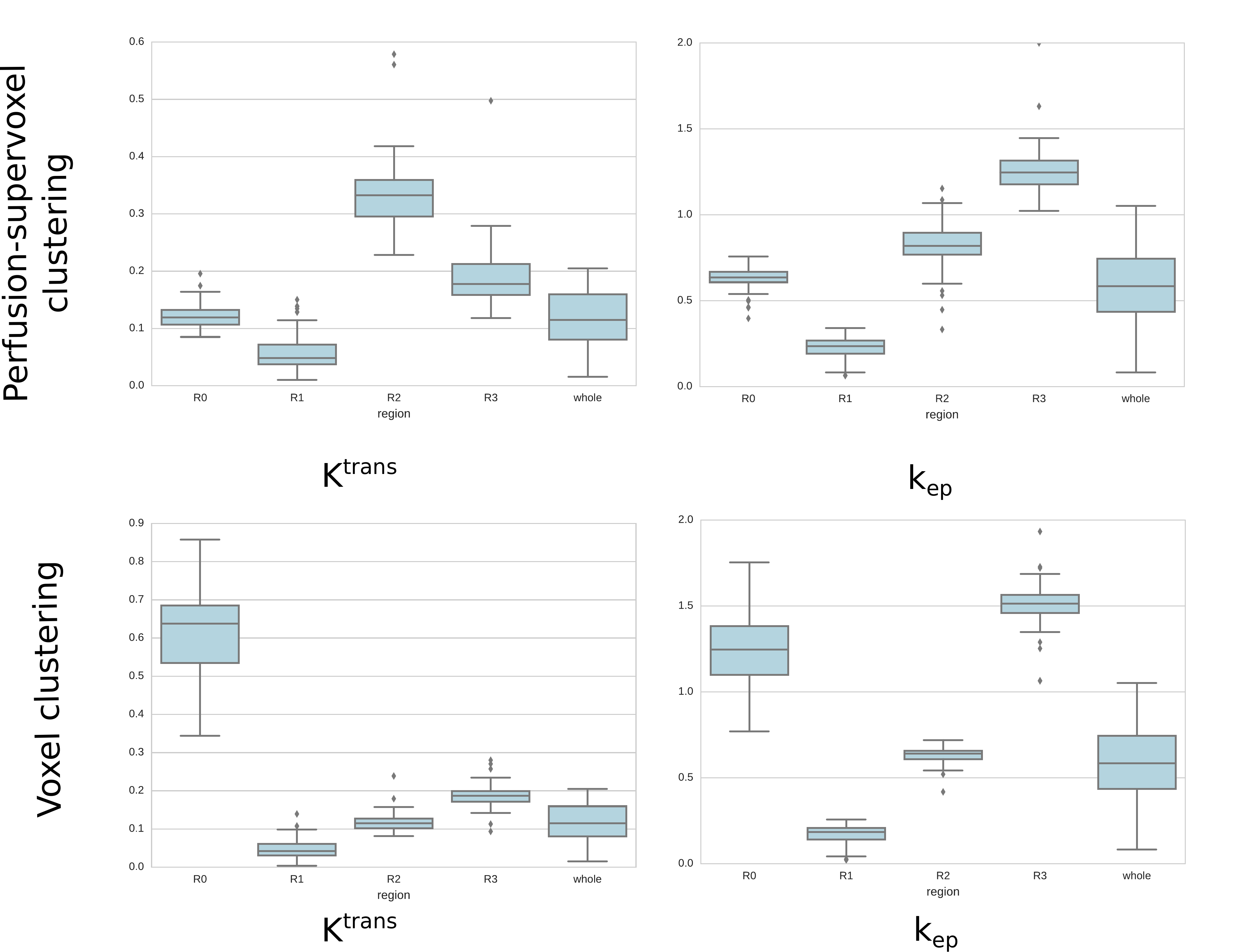}
\caption{Mean $K^{trans}$ and $k_{ep}$ for each labelled region and the whole tumour}
\label{fig:regpropmean}
\end{figure*}

Figure \ref{fig:regpropmean} shows boxplots of each of the four labelled regions (and the mean tumour values) for the entire dataset for both supervoxel and voxelwise clustering. Clustering without a supervoxel representation leads to potential outlier clusters (R0) that cover a large range of parameter values and don't have any regional significance. These outliers are not present during clustering of supervoxels, and instead we can capture more interesting parameter relationships, such as regions that show apparent decoupling of the perfusion parameter maps, i.e. some regions exhibit show [low $k_{ep}$ low $K^{trans}$], and [high $k_{ep}$, high $K^{trans}$] -- see regions R0, R1 and R2 -- while others exhibit [high $K^{trans}$, low $k_{ep}$] (region R3), which could have significance for identifying regions with different biological properties.

\section{Discussion and conclusion}

Supervoxel analysis has a wide range of applications in computer vision and medical image analysis because of the ability to reduce an image into a set of meaningful subregions. We have demonstrated that, with three modifications, the standard SLIC method can be made more effective for analysis within defined regions such as a tumour or organ. We demonstrate improved invariance and quality of the subregions on a number of examples, and show that \emph{mask}SLIC provides more meaningful subregions on the BRATS brain tumour segmentation challenge. Finally, we demonstrate that \emph{mask}SLIC with clustering is a very promising technique for developing stable tumour subregions for a dataset of cases but needs further validation on histology. 

A limitation of this method is that the distance transform based point placement is slower than a grid point placement. Time is saved by only performing the method within a defined region so the speed of the method compared to SLIC depends on a trade-off between the size of the irregular region and the original region. 

\subsubsection*{Resources}
A demo of the method is available as \href{http://maskslic.birving.com}{\underline{http://maskslic.birving.com}} and example code is available at \href{https://github.com/benjaminirving/maskSLIC}{\underline{https://github.com/benjaminirving/maskSLIC}}.

\section{Acknowledgements}
We would like to thank the CRUK/EPSRC Oxford Cancer Imaging Centre for supporting this work. IP acknowledges the support of RCUK Digital Economy Programme (grant number EP/G036861/1), Oxford Centre for Doctoral Training in Healthcare Innovation. 

\bibliographystyle{IEEEtran}
\bibliography{library.bib}

\end{document}